%% file: main.tex
\pdfoutput=1

\documentclass[11pt]{article}

\usepackage{emnlp2021}

\usepackage{times}
\usepackage{latexsym}

\usepackage[T1]{fontenc}

\usepackage[utf8]{inputenc}

\usepackage{microtype}

\usepackage{graphicx}
\usepackage{placeins}
\usepackage{xcolor}
\usepackage{comment}
\usepackage{todonotes}
\newcommand{\best}[1]{\textbf{#1}}
\newcommand{\relation}[1]{\textsc{#1}}

\renewcommand{\cite}[1]{\citep{#1}}
\newcommand{\kgi}[1]{$\text{KGI}_{#1}$}

\usepackage{amsmath}

\DeclareMathOperator*{\argmax}{argmax}

\title{Robust Retrieval Augmented Generation for Zero-shot Slot Filling}

\author{Michael Glass, Gaetano Rossiello, Md Faisal Mahbub Chowdhury, Alfio Gliozzo\\
        IBM Research AI \\ Thomas J. Watson Research Center, NY}
        
\begin{document}
\maketitle

\begin{abstract}
Automatically inducing high quality knowledge graphs from a given collection of documents still remains a challenging problem in AI. One way to make headway for this problem is through advancements in a related task known as slot filling. In this task, given an entity query in form of \textsc{[Entity, Slot, ?]}, a system is asked to `fill' the slot by generating or extracting the missing value exploiting evidence extracted from relevant passage(s) in the given document collection.
The recent works in the field try to solve this task in an end-to-end fashion using retrieval-based language models. 
In this paper, we present a novel approach to zero-shot slot filling that extends dense passage retrieval with hard negatives and robust training procedures for retrieval augmented generation models.
Our model reports large improvements on both T-REx and zsRE slot filling datasets, improving both passage retrieval and slot value generation, and ranking at the top-1 position in the KILT leaderboard. 
Moreover, we demonstrate the robustness of our system showing its domain adaptation capability on a new variant of the TACRED dataset for slot filling, through a combination of zero/few-shot learning.
We release the source code and pre-trained models\footnote{Our source code is available at: \url{https://github.com/IBM/kgi-slot-filling}}.
\end{abstract}

\section{Introduction}
\begin{figure}[t]
  \centering
  \includegraphics[width=1\linewidth]{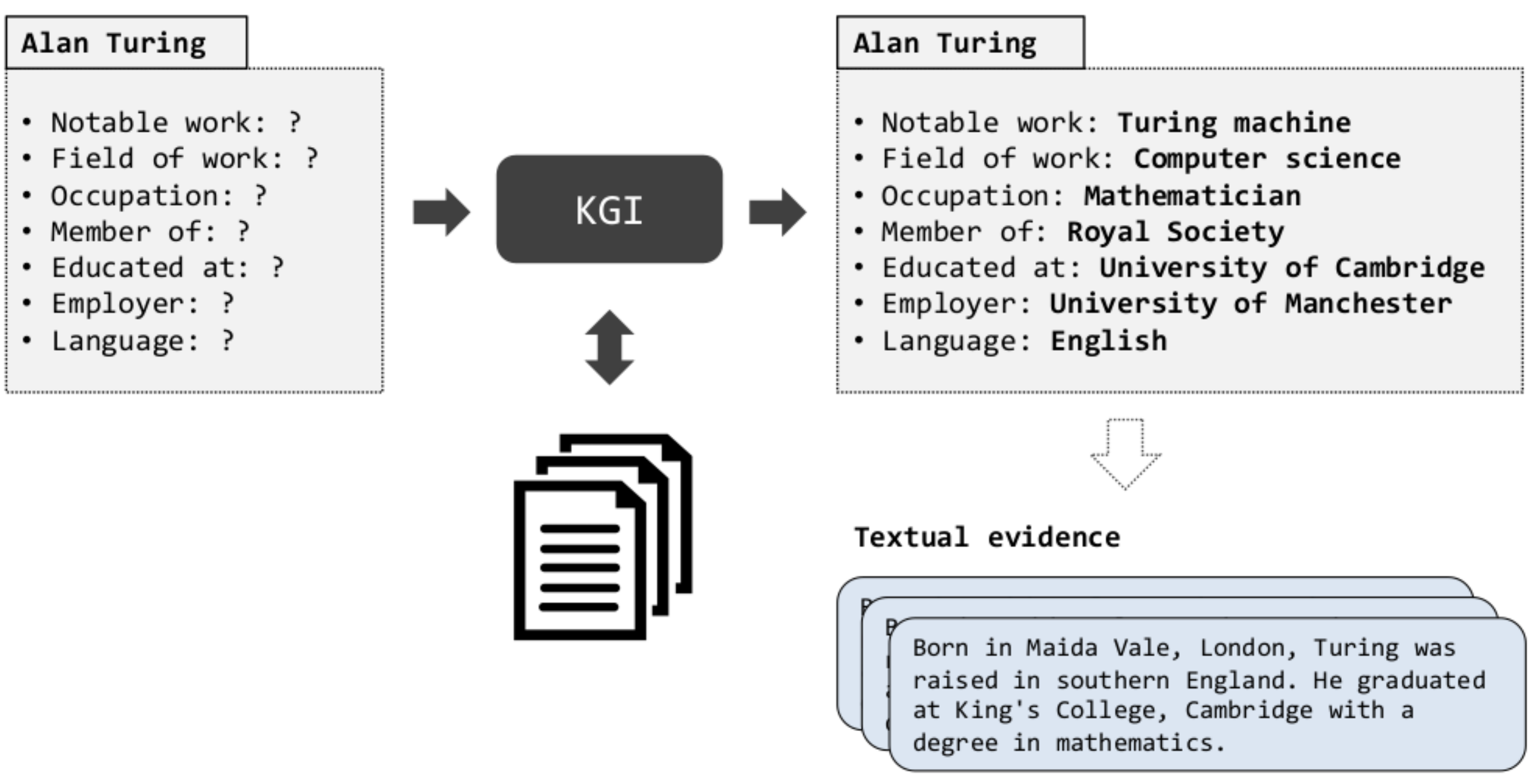}
  \caption{Slot Filling task}
  \label{fig.task}
\end{figure}

Slot filling is a sub-task of Knowledge Base Population (KBP), where the goal is to recognize a pre-determined set of relations for a given entity and use them to populate infobox like structures. This can be done by exploring the occurrences of the input entity in the corpus and gathering information about its slot fillers from the context in which it is located. A slot filling system processes and indexes a corpus of documents. 
Then, when prompted with an entity and a number of relations, it fills out an infobox for the entity. Some slot filling systems provide evidence text to explain the predictions.
Figure \ref{fig.task} illustrates the slot filling task. 



Many KBP systems described in the literature commonly involve complex pipelines for named entity recognition, entity co-reference resolution and relation extraction \cite{DBLP:conf/tac/EllisGFKSBS15}. 
In particular, the task of extracting relations between entities from text has been shown to be the weakest component of the chain. 
The community proposed different solutions to improve relation extraction performance, such as rule-based \cite{DBLP:conf/acl/AngeliPM15}, supervised \cite{DBLP:conf/emnlp/ZhangZCAM17}, or distantly supervised \cite{DBLP:conf/semweb/GlassGHMR18}. However, all these approaches require a considerable human effort in creating hand-crafted rules, annotating training data, or building well-curated datasets for bootstrapping relation classifiers.


Recently, pre-trained language models have been used for slot filling~\cite{DBLP:conf/akbc/PetroniLPRWM020}, opening a new research direction that might provide an effective solution to the aforementioned problems. 
In particular, the KILT benchmark~\cite{kilt}, standardizes two zero-shot slot filling tasks, zsRE~\cite{zsre} and T-REx~\cite{trex}, providing a competitive evaluation framework to drive advancements in slot filling. However, the best performance achieved by the current retrieval-based models on the two slot filling tasks in KILT are still not satisfactory. This is mainly due to the lack of retrieval performance that affects the generation of the filler as well.

In this work, we propose \kgi{} (Knowledge Graph Induction), a robust system for slot filling based on advanced training strategies for both Dense Passage Retrieval (DPR) and Retrieval Augmented Generation (RAG) that
shows large gains on both T-REx (+38.24\% KILT-F1) and zsRE (+21.25\% KILT-F1) datasets if compared to previously submitted systems.
We extend the training strategies of DPR with \emph{hard negative mining} \cite{hardNegative}, demonstrating its importance in training the context encoder.

In addition, we explore the idea of adapting \kgi{} to a new domain.
The domain adaptation process consists of indexing the new corpus using our pre-trained DPR and substituting it in place of the original Wikipedia index.
This enables zero-shot slot filling on the new dataset with respect to a new schema, avoiding the additional effort needed to re-build NLP pipelines. We provide a few additional examples for each new relation, showing that zero-shot performance quickly improves with a few-shot learning setup.  
We explore this approach on a variant of the TACRED dataset~\cite{tacred_revisited} that we specifically introduce to evaluate the zero/few-shot slot filling task for domain adaption.

The contributions of this work are as follows:
\begin{enumerate}
    \item We describe an end-to-end solution for slot filling, called \kgi{}, that improves the state-of-the-art in the KILT slot filling benchmarks by a large margin.
    \item We demonstrate the effectiveness of hard negative mining for DPR when combined with end-to-end training for slot filling tasks.
    \item We evaluate the domain adaptation of \kgi{} using zero/few-shot slot filling, demonstrating its robustness on zero-shot TACRED, a benchmark released with this paper.
    \item We publicly release the pre-trained models and source code of the \kgi{} system.
\end{enumerate}


Section \ref{sec.related} present an overview of the state of the art in slot filling. Section \ref{sec.approach} describes our \kgi{} system, providing details on the DPR and RAG models and describing our novel approach to hard negatives.
Our system is evaluated in Sections \ref{sec.eval_kilt} and \ref{sec.eval_tacred} which include a detailed analysis. Section \ref{sec.conclusion} concludes the paper and highlights some interesting direction for future work.

\section{Related Work}\label{sec.related}
The use of language models as sources of knowledge \cite{DBLP:conf/emnlp/PetroniRRLBWM19, DBLP:conf/emnlp/RobertsRS20, DBLP:journals/corr/abs-2010-11967, DBLP:conf/akbc/PetroniLPRWM020}, has opened tasks such as zero-shot slot filling to pre-trained transformers. 
Furthermore, the introduction of retrieval augmented language models such as RAG \cite{rag} and REALM \cite{realm} also permit providing textual provenance for the generated slot fillers. 



KILT \cite{kilt} was introduced with a number of baseline approaches. The best performing of these is RAG \cite{rag}. The model incorporates DPR \cite{dpr} to first gather evidence passages for the query, then uses a model initialized from BART \cite{bart} to do sequence-to-sequence generation from each evidence passage concatenated with the query in order to generate the answer. In the baseline RAG approach only the query encoder and generation component are fine-tuned on the task. The passage encoder, trained on Natural Questions \cite{naturalquestions} is held fixed.
Interestingly, while it gives the best performance of the baselines tested on the task of producing slot fillers, its performance on the retrieval metrics is worse than BM25 \cite{kilt}. This suggests that fine-tuning the entire retrieval component could be beneficial. 
Another baseline in KILT is BART$_{LARGE}$ fine-tuned on the slot filling tasks but without the usage of the retrieval model.

In an effort to improve the retrieval performance, Multi-task DPR \cite{multidpr} used the multi-task training of the KILT suite of benchmarks to train the DPR passage and query encoder.  The top-3 passages returned by the resulting passage index were then combined into a single sequence with the query and a BART model was used to produce the answer.  This resulted in large gains in retrieval performance.

DensePhrases \cite{densephrases} is a different approach to knowledge intensive tasks with a short answer. Rather than index passages which are then consumed by a reader or generator component, it indexes the phrases in the corpus that can be potential answers to questions, or fillers for slots. Each phrase is represented by the pair of its start and end token vectors from the final layer of a transformer initialized from SpanBERT~\cite{spanbert}.


GENRE \cite{genre} addresses the retrieval task in KILT slot filling by using a sequence-to-sequence transformer to generate the title of the Wikipedia page where the answer can be found.  This method can produce excellent scores for retrieval but it does not address the problem of producing the slot filler.
It is trained on BLINK \cite{blink} and all KILT tasks jointly.

Open Retrieval Question Answering (ORQA) \cite{lee-etal-2019-latent} introduced neural information retrieval for the related task of factoid question answering. Like DPR, the retrieval is based on a bi-encoder BERT \cite{bert} model. Unlike DPR, ORQA projects the BERT {\small [CLS]} vector to a lower dimensional (128) space. It also uses the \emph{inverse cloze} pre-training task for retrieval, while DPR does not use retrieval specific pre-training.

\section{Knowledge Graph Induction}\label{sec.approach}

\begin{figure}[h!]
   \centering
   \includegraphics[width=0.9\linewidth]{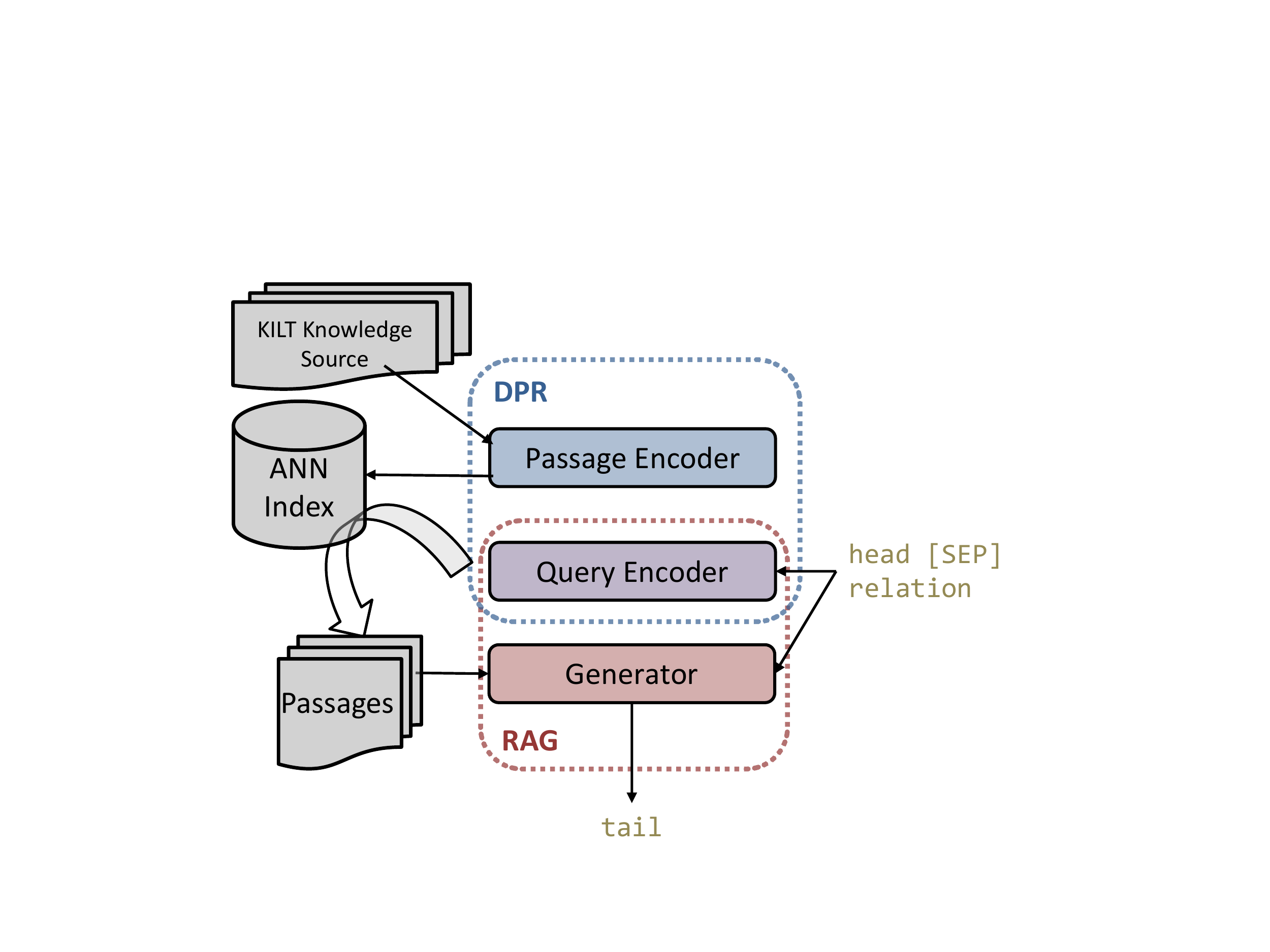}
   \caption{\kgi{} Architecture}
   \label{fig.arch}
\end{figure}

Figure \ref{fig.arch} shows \kgi{}, 
our approach to zero-shot slot filling, combining a DPR model and RAG model, both trained for slot filling.  
We initialize our models from the Natural Questions~\cite{naturalquestions} trained models for DPR and RAG available from Hugging Face \cite{huggingface}\footnote{\url{https://github.com/huggingface/transformers}}.
We then employ a two phase training procedure: first we train the DPR model, i.e. both the query and context encoder, using the KILT provenance ground truth. Then we train the sequence-to-sequence generation and further train the query encoder using only the target tail entity as the objective. It is important to note that the same query encoder component is trained in both phases. 

\subsection{DPR for Slot Filling}

Our approach to DPR training for slot filling is an adaptation of the question answering training in the original DPR work \cite{dpr}.
We first index the passages using a traditional keyword search engine, Anserini\footnote{\url{https://github.com/castorini/anserini}}. The head entity and the relation are used as a keyword query to find the top-k passages by BM25. Passages with overlapping paragraphs to the ground truth are excluded as well as passages that contain a correct answer. The remaining top ranked result is used as a hard negative for DPR training. 
This is the hard negative mining strategy used by DPR \cite{dpr} and Multi-DPR \cite{multidpr}.

\begin{figure}[thb]
   \centering
   \includegraphics[width=0.9\linewidth]{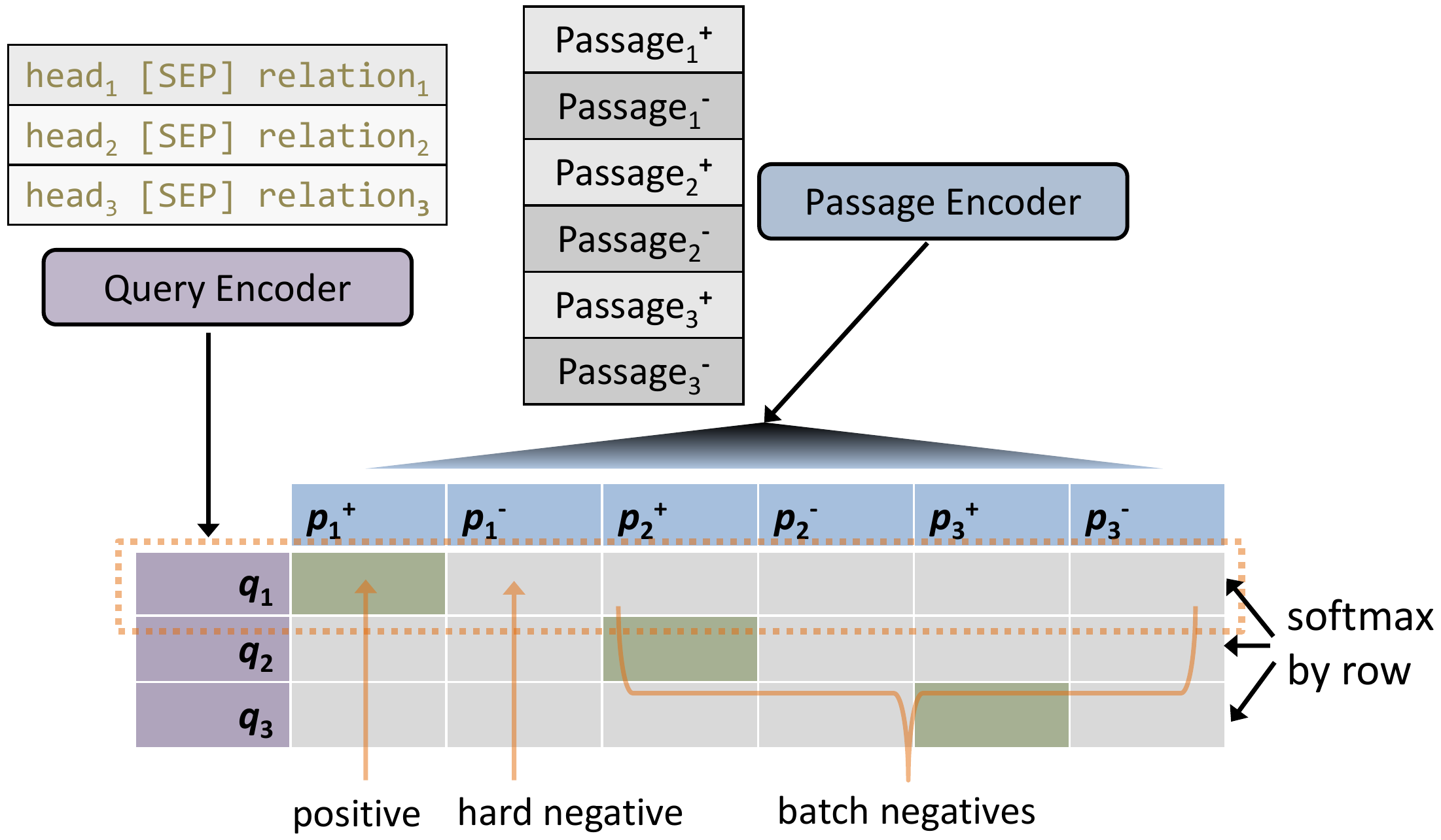}
   \caption{DPR Training}
   \label{fig.dpr}
\end{figure}

After locating a hard negative for each query, the DPR training data is a set of triples: query, positive passage (given by the KILT ground truth provenance) and the hard negative passage. Figure \ref{fig.dpr} shows the training process for DPR. For each batch of training triples, we encode the queries and passages independently. The passage and query encoders are BERT \cite{bert} models. Then we find the inner product of all queries with all passages. 
The negatives for a given query are therefore the hard negative and the batch negatives, i.e. the positive and hard negative passages for other queries in the batch. 
After applying a softmax to the score vector for each query, the loss is the negative log-likelihood for the positive passages.

Using the trained DPR passage encoder we generate vectors for the approximately 32 million passages in our segmentation of the KILT knowledge source. Though this is a computationally expensive step, it is easily parallelized.  
The passage-vectors are then indexed with an ANN (Approximate Nearest Neighbors) data structure, in this case HNSW (Hierarchical Navigable Small World)\cite{hnsw} using the open source FAISS library \cite{faiss}\footnote{\url{https://github.com/facebookresearch/faiss}}.
We use scalar quantization down to 8 bits to reduce the memory size.


The query encoder is also trained for slot filling alongside the passage encoder. We inject the trained query encoder into the RAG model for Natural Questions. Due to the loose coupling between the query encoder and the sequence-to-sequence generation of RAG, we can update the pre-trained model's query encoder without disrupting the quality of the generation.

Unlike previous work on zero-shot slot filling, we are training the DPR model specifically for the slot filling task. In contrast, the RAG baseline \cite{kilt} used DPR pre-trained on Natural Questions, and Multi-DPR \cite{multidpr} trained on all KILT tasks jointly.

\subsection{RAG for Slot Filling}
\begin{figure}[thb]
   \centering
   \includegraphics[width=0.9\linewidth]{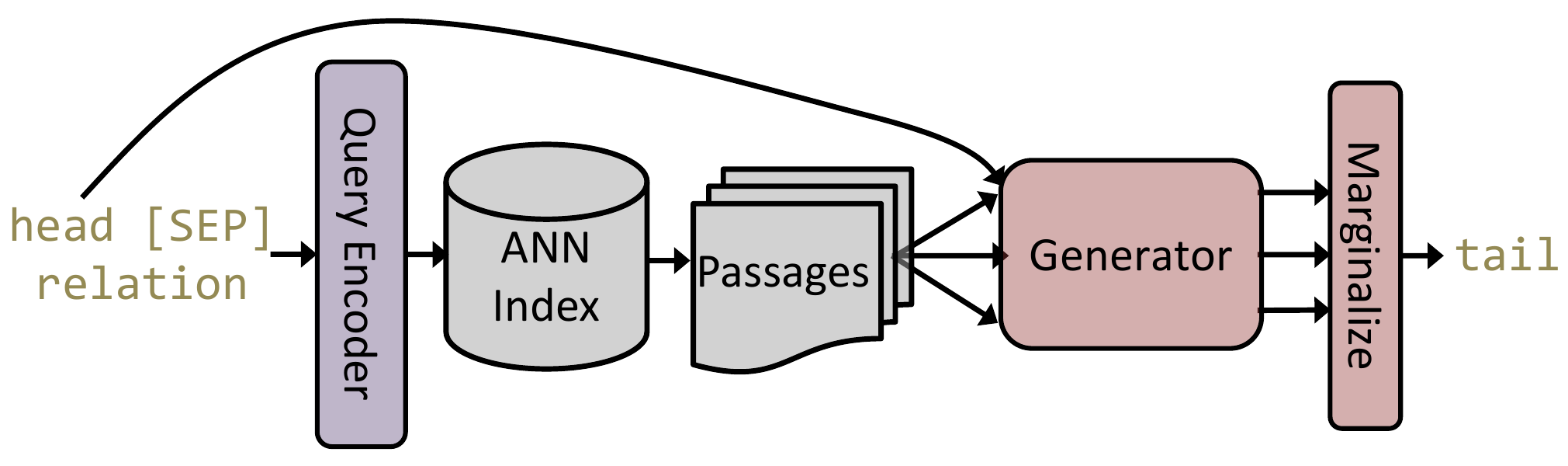}
   \caption{RAG Architecture}
   \label{fig.rag}
\end{figure}

Figure \ref{fig.rag} illustrates the architecture of RAG \cite{rag}.
The RAG model is trained to predict the ground truth tail entity from the head and relation query.  First the query is encoded to a vector and the top-k (we use $k=5$) relevant passages are retrieved from the ANN index.  The query is concatenated to each passage and the generator predicts a probability distribution over the possible next tokens for each sequence. These predictions are weighted according to the score between the query and passage - the inner product of the query vector and passage vector.   

Marginalization then combines the weighted probability distributions to give a single probability distribution for the next token.
This enables RAG to train the query encoder through its impact in generation, learning to give higher weight to passages that contribute to generating the correct tokens.  
Formally, the inputs to the BART model are sequences ($s_j = p_j\ \text{\small [SEP]}\ q$) that comprise a query $q$ plus retrieved passage $p_j$. The probability for each sequence is determined from the softmax over the retrieval scores ($\mathbf{z_r}$) for the passages. The probability for each output token $t_i$ given the sequence $s_j$ is a softmax over BART's token prediction logits.  Therefore the total probability for each token $t_i$ is the log-likelihood summed over all sequences, weighted by each sequence's probability.

\begin{align*}
    P(s_j) & = softmax(\mathbf{z_r})_j\\
    P(t_i | s_j) & = softmax(\text{BART}(s_j)_i)_{t_i} \\
    P(t_i)  & = \sum_{j} P(t_i | s_j) \cdot P(s_j)
\end{align*}

Beam search is used at inference time to select the overall most likely tail entity. This is the standard beam search for natural language generation in deep neural networks \cite{dnnBeam}, the only difference is in the way the next-token probabilities are obtained.

\input{dns.tex}


\section{KILT Experiments}
\label{sec.eval_kilt}
\input{kilt_experiments.tex}

\subsection{Analysis}

\input{kilt_analysis2.tex}


\section{Domain Adaptation Experiments}\label{sec.eval_tacred}
In this section, we evaluate the domain adaptation capability of \kgi{}. For this purpose, we re-organize a dataset specifically designed to evaluate standard supervised relation extraction models, such as TACRED, with the aim to create a zero-shot (and few-shot) slot filling benchmark where the documents are written with a different style than Wikipedia, and the relations in the KG are different from those in Wikidata. In order to perform an in-depth comparison and analysis, we also propose a new set of ranking baselines and use metrics which are suitable to better evaluate the slot filling task in a zero-shot setup.

\subsection{Zero-shot TACRED}\label{sec.tacred_dataset}
\begin{table*}[t]
    \centering
    \small
    \begingroup
    \renewcommand{\arraystretch}{1.2} 
    \begin{tabular}{l | l | l | p{0.34\linewidth}}
      \textbf{Subject} & \textbf{Relation/Slot} & \textbf{Object} & \textbf{Passage} \\ \hline\hline
      
      Dominick Dunne & per:employee\_of & Vanity Fair & \tiny Dominick Dunne, the author, television personality and Vanity Fair reporter who covered the trials of socialite Claus von Bulow. \\
      
      Dominick Dunne & per:age & 83 & \tiny Dominick Dunne, 83, a crime story author, in New York City. \\
      Dominick Dunne & per:siblings & John Gregory Dunne & \tiny Dominick Dunne, author of crime stories dies Born in 1925 in Hartford, Connecticut, was part of a famous family that included his brother, novelist and screenwriter John Gregory Dunne. \\\hline
      
      ALICO & org:country\_of\_headquarters & US & \tiny AIA says IPO raised 205 billion US dollars AIG said Monday it had also raised 162 billion dollars by selling unit American Life Insurance Company (ALICO) to MetLife Inc. \\ 
      
      ALICO & org:top\_members & Christopher J. Swift & \tiny Alico's chief financial officer, Christopher J. Swift, added that the bonds were issued by companies in many commercial sectors, which diversified the portfolio. \\
      
      ALICO & org:parents & AIG & \tiny AIG said it had transferred ownership to the Federal Reserve Bank of parts of two subsidiaries, ALICO which is active in life assurance in the United States and AIA which provides life assurance abroad. \\\hline
      
    \end{tabular}
    \endgroup
    \caption{Examples of annotations from TACRED dataset for both \textit{person} and \textit{organization} infoboxes.}
    \label{tab:tacred}
\end{table*}

The TACRED dataset was originally proposed by~\citet{DBLP:conf/emnlp/ZhangZCAM17} with the goal to provide a high-quality training set to supervise a relation extraction model which is shown to be competitive on TAC-KBP 2015~\cite{DBLP:conf/tac/EllisGFKSBS15}. 
The target KG schema consists of two infoboxes modeling the \textit{person} and \textit{organization} entity types, with 41 relation types in total.
For our experiments, we adopt a revisited version of TACRED \cite{tacred_revisited}, in which a second stage crowdsourcing is performed to further improve the quality of the annotations and resolve conflicts among relations.

In a typical supervised relation extraction setup, a model is trained to predict (i.e. classify) the right relation type given a textual passage and two entity mentions as inputs.
In this paper we used the TACRED dataset as a slot filling benchmark, using the following procedure: 1) we first create the corpus by merging all the plain textual passages from the instances in the train, dev and test sets; 2) we collect the annotated triples, i.e. subject-relation-object, from the test data to come up with a ground-truth KG to be used for slot filling evaluation\footnote{$79.5\%$ of the overall instances are labeled as no relation. We exclude these instances from the ground truth KG, but we retain them in the textual corpus.}; 3) we remove all the triples from the original test set where the subjects are pronouns. 
The resulting KG consists of 2673 slot filling \emph{test} instances.
Similarly, we acquire a KG from the train/dev sets to further fine-tune the \kgi{} system as described in the next section.
To enable zero-shot experiments, we also convert each relation label into a relation phrase by removing the namespaces \textit{per:} and \textit{org:}, and replacing the `\_' character with a space.
Finally, for each pre-annotated entity in the corpus, we pre-compute an inverted index consisting of a list of co-occurring entities in the textual passages. We use this inverted index to compare our model with a set of ranking baselines.

An example of the obtained ground truth is illustrated in Table~\ref{tab:tacred}: given the query \textit{[Dominick Dunne, employee of, ?]}, a slot filling system is supposed to identify the missing slot with \textit{Vanity Fair}, i.e. the gold standard object in the KG, by retrieving it from the collection of passages.

\subsection{Slot Filling Evaluation}
\paragraph{Task}
Given a slot filling query $(e, s, ?)$ and a list of possible slot values $[v_{1}, ..., v_{n}]$, where $e$ is the entity as subject, $s$ is the slot/relation and $v_{i}$ are the object candidates that co-occur with $e$ in the corpus, we can frame the zero-shot slot filling as a ranking problem: $\argmax_i{score_M(e, s, v_{i})}$. $score_M$ is a function that takes as input a triple and provide a score based on the model $M$. Turning the slot filling into a ranking problem has two advantages: 1) we can compare the generative approach with a new set of baselines, and 2) we can limit the generation of the slot values to a pre-defined set of domain specific entities.


\paragraph{Models}
In order to adapt  \kgi{1}, as pre-trained on T-REx, to the TACRED corpus, we indexed the textual passages using DPR, as described in Section~\ref{sec.approach}. Then we replaced the original Wikipedia index with this new index. During the inference step, we restrict the generation of the slot values using the list of object candidates, i.e. the entities which co-occur with the subject from the inverted index, to facilitate comparability to a set of ranking baselines.
To this aim, we adopt the technique described by~\citet{genre} to restrict the vocabulary of tokens during the generation.

We use three baselines to compare with our approach for this zero-shot slot filling task.
\textbf{PMI} is implemented using the pointwise mutual information between $e$ and $v_{i}$ based on their co-occurrence in the corpus.
Also, we train a \textbf{Word2Vec}~\cite{DBLP:conf/nips/MikolovSCCD13} skip-gram model on the textual corpus, and we use it to implement the scoring function as $cosine(e + s, v_{i})$, for each candidate filler $v_{i}$. It is based on the assumption that a relation $s$ between two (multi)word embeddings $e$ and $v$ can be represented as an offset vector $(v - e) = s \iff (e + s) = v$ ~\cite{DBLP:conf/naacl/RossielloGFFG19,DBLP:conf/acl/VylomovaRCB16}. 
Finally, \textbf{GPT-2} computes the perplexity of the fragment of text by concatenating the tokens in $e$, $s$ and each $v_{i}$~\cite{radford2019language}.


\begin{table}[t]
    \small
    \centering
    \begingroup
    \renewcommand{\arraystretch}{1.2} 
    \begin{tabular}{r|c|c|c|c}
          & \textbf{MRR} & \textbf{HIT@1} & \textbf{HIT@5} & \textbf{HIT@10} \\\hline\hline
         PMI & 20.20 & 10.89 & 26.49 & 37.30 \\
         Word2Vec  & 25.24 & 13.83 & 34.92 & 47.60 \\
         GPT-2  & 17.62 & 8.37 & 23.72 & 35.34 \\
         
         \kgi{1} 0-shot  & \best{43.98} & \best{28.51} & \best{64.31} &  \best{76.06}  \\ \hline \hline
         \kgi{1} 1-shot  & 48.86 & 33.89 & 66.63	 &  78.75\\
         \kgi{1} 4-shot & \best{\textit{53.28}} & \best{\textit{38.8}} & \best{\textit{70.45}} & \best{\textit{79.35}} \\ \hline
         
         

    \end{tabular}
    \endgroup
    \caption{Zero/few-shots results on TACRED}
    \label{tab:zero-shot}
\end{table}

\paragraph{Metrics}
Due to the similarity of slot filling with the knowledge base completion task, we use Mean Reciprocal Rank (MRR) and HIT@k, with k = [1, 5, 10], as evaluation metrics~\cite{DBLP:conf/nips/BordesUGWY13}. 
Note that HIT@1 has the same meaning of the accuracy for the downstream task on KILT.

\paragraph{Results}
Table~\ref{tab:zero-shot} reports the results of our evaluation. \kgi{1} achieves substantially better performance than the aforementioned zero-shot baselines on all evaluation metrics. However, HIT@1 is $\sim28\%$ which is significantly lower compared with the numbers reported on the datasets in KILT. This begs the question, how to further improve the transfer learning capabilities of these generative models?
Interestingly, HIT@5/10 are high (i.e. $\sim64\%/76\%$). This indicates our approach would be useful in a human-in-the-loop scenario by providing valuable candidates for the fillers that can be further validated.

\begin{table}[t]
\small
\begin{center}
\begin{tabular}{rrr}
\textbf{Hyperparameter} & \textbf{RAG} \\
\hline
learn rate & 3e-6 \\
batch size & 1 \\
epochs & 3 \\
warmup instances & 0 \\
learning schedule & linear \\
max grad norm & 1  \\
weight decay & 0  \\
Adam epsilon & 1e-8
\end{tabular}
\end{center}
\caption{\kgi{1} hyperparameters for TACRED few-shot}
\label{tbl.tacred_hypers}
\end{table}

For this purpose, we also conduct few-shot experiments to understand the robustness of \kgi{1} by fine-tuning it with very limited amounts of training examples.
We randomly pick \texttt{n} example(s) for each relation type from the TACRED training set, with n = [1, 4]. 
Table \ref{tbl.tacred_hypers} gives our hyperparameters for the TACRED few-shot experiments.
We show that our system benefits from additional domain specific training data selected from TACRED. 
Just using one example and four examples per relation, HIT@1 improves $\sim5$ and $\sim10$ percentage points respectively.





\section{Conclusion}\label{sec.conclusion}
In this paper, we presented \kgi{}, a novel approach to zero-shot slot filling. \kgi{} improves Dense Passage Retrieval using hard negatives from the dense index, and implements a robust training procedures for Retrieval Augmented Generation.
We evaluated \kgi{}  on both T-REx and zsRE slot filling datasets,  ranking at top-1 position in the KILT leaderboard with a net improvement of +38.24 and +21.25 percentage points in KILT-F1, respectively.  
Moreover, we proposed and release a new benchmark for zero/few-shot slot filling based on TACRED to evaluate domain adaptation where our system obtained much better zero-shot results compared with the baselines. 
In addition, we have observed significant improvement in results for \kgi{} when rapidly fine-tuned in a few-shot setting. 
This work opens promising future research directions for slot filling and other related tasks. We plan to apply DPR with dense negative sampling to other tasks in the KILT benchmark, including dialogue and question answering. Likewise, an in-depth investigation on more effective strategies for domain adaptation, such as the combination of zero-shot and few-shot learning involving human-in-the-loop techniques, would be another interesting direction to explore. 


\FloatBarrier

\bibliography{anthology,custom}
\bibliographystyle{acl_natbib}


\end{document}

%% file: dns.tex
\subsection{Dense Negative Sampling}

As Figure \ref{fig.arch} shows, the DPR question encoder is trained both by DPR and later by RAG. To examine the influence of this additional training from RAG on the retrieval performance, we compare retrieval metrics before and after RAG fine-tuning.  Table \ref{tbl.retrieval} shows the large gains from training with RAG after DPR. 
Note that RAG training is using the weak supervision of the passage's impact in producing the correct answer, rather than the ground truth provenance of DPR training. Since this is likely a disadvantage, we explore the other key difference with DPR and RAG training: RAG uses negatives drawn from the trained index rather than from BM25.

\begin{table}[thb]
\small
\begin{center}
\begingroup
\renewcommand{\arraystretch}{1.2} 
\resizebox{\linewidth}{!}{%
\begin{tabular}{r|cc|cc}
 & \multicolumn{2}{c|}{\textbf{T-REx}} & \multicolumn{2}{c}{\textbf{zsRE}}\\\hline
& \textbf{R-Prec}  & \textbf{R@5} & \textbf{R-Prec}  & \textbf{R@5} \\
\hline
DPR$_{NQ}$ & 19.50 & 29.80 & 45.49 & 60.77 \\
DPR$_{NQ}$+RAG & 53.04 & 65.54 & 68.13 & 79.19 \\
DPR$_{BM25}$ & 49.02 & 63.34 & 94.55 & 98.17 \\
DPR$_{BM25}$+RAG & 65.02 & 75.52 & 96.89 & 98.01 \\
DPR$_{DNS}$ & \textit{42.62} & \textit{55.09} & 97.53 &	99.30 \\
DPR$_{DNS}$+RAG & \best{74.34} & \best{82.89} & \best{98.60} & \best{99.70}
\end{tabular}
}
\endgroup
\end{center}
\caption{\label{tbl.retrieval}Analysis of retrieval by DPR and RAG on Dev sets}
\end{table}

To replicate this feature of RAG in DPR, we introduce hard negatives mined from the learned index. Using the KILT trained DPR models, we index the passages. Then we gather hard negatives for DPR training, with one difference: rather than locating the hard negative passages by BM25, we find the passage by ANN search over the learned dense vector index.  We train for an additional two epochs using these hard negatives. Table \ref{tbl.retrieval} shows the performance of the different approaches to retrieval. DPR$_{NQ}$ is the DPR model pre-trained on Natural Questions. DPR$_{BM25}$ further trains DPR$_{NQ}$ on the KILT data with BM25 hard negatives. Rows with +RAG further train the question encoder through RAG.  The row DPR$_{DNS}$ (\textbf{D}ense \textbf{N}egative \textbf{S}ampling) shows the performance of retrieval immediately after DNS training. Surprisingly, this results in lower performance for T-REx relative to DPR$_{BM25}$. However, further training the DNS model with RAG results in our best performance for both T-REx and zsRE.
Since RAG does not update the context encoder, DNS training is the only training for the context encoder when negatives are drawn from the dense vector index.

After training with DNS the FAISS indexing with scalar quantization becomes prohibitively slow.  We therefore remove all quantization and use four shards (the index is split into four, with the results of each query merged) for our experiments with DNS enabled \kgi{}.

%% file: kilt_experiments.tex
Table \ref{tbl.datasets} gives statistics on the two zero-shot slot filling datasets in KILT. While the T-REx dataset is larger by far in the number of instances, the training sets have a similar number of distinct relations. We use only 500k training instances of T-REx in our experiments to increase the speed of experimentation.

\begin{table}
\begin{center}
\begingroup
\renewcommand{\arraystretch}{1.2} 
\resizebox{\linewidth}{!}{%
\begin{tabular}{r|ccc|ccc}
 & \multicolumn{3}{c|}{\textbf{Instances}} & \multicolumn{3}{c}{\textbf{Relations}}\\ \hline
\textbf{Dataset} & \textbf{Train}  & \textbf{Dev} & \textbf{Test} & \textbf{Train}  & \textbf{Dev} & \textbf{Test}\\
\hline
zsRE  &	148K & 3724 & 4966 &	84 & 12 & 24 \\
T-REx &	2284K & 5000 & 5000 &	106 & 104 & 104
\end{tabular}
}
\endgroup
\end{center}
\caption{\label{tbl.datasets}Slot filling datasets in KILT}
\end{table}

Since the transformers for passage encoding and generation can accept a limited sequence length, we segment the documents of the KILT knowledge source (2019/08/01 Wikipedia snapshot) into passages. The ground truth provenance for the slot filling tasks is at the granularity of paragraphs, so we align our passage segmentation on paragraph boundaries when possible.  If two or more paragraphs are short enough to be combined, we combine them into a single passage and if a single paragraph is too long, we truncate it.

\subsection{\kgi{} Hyperparameters}
\begin{table}[t]
\small
\begin{center}
\begin{tabular}{rrr}
\textbf{Hyperparameter} & \textbf{DPR}  & \textbf{RAG} \\
\hline
learn rate  & 5e-5 & 3e-5 \\
batch size & 128 & 128 \\
epochs & 2 & 1\\
warmup instances & 0 & 10000 \\
learning schedule & linear & triangular \\
max grad norm & 1 & 1 \\
weight decay & 0 & 0 \\
Adam epsilon & 1e-8 & 1e-8
\end{tabular}
\end{center}
\caption{\kgi{} hyperparameters}
\label{tbl.hypers}
\end{table}

We have not done hyperparameter tuning, instead using hyperparameters similar to the original works on training DPR and RAG. Table \ref{tbl.hypers} shows the hyperparameters used in our experiments. 
We train our models on T-REx using only the first 500k instances.
For \kgi{1} we use the same hyperparameters except that zsRE is trained for two epochs. In both \kgi{} systems we use the default of five passages retrieved for each query for use in RAG.

\subsection{Model Details}

\paragraph{Number of parameters}
\kgi{} is based on RAG and has the same number of parameters: $2 \times 110M$ for the BERT$_{BASE}$ query and passage encoders and $400M$ for the BART$_{LARGE}$ sequence-to-sequence generation component: $620M$ in total.

\paragraph{Computing infrastructure}
Using a single NVIDIA V100 GPU DPR training of two epochs takes approximately 24 hours for T-REx and 2 hours for zsRE.
Using a single NVIDIA P100 GPU RAG training for 500k T-REx instances takes two days and 147k instances of zsRE takes 15 hours.
The FAISS index on the KILT knowledge source requires a machine with large memory, we use 256GB memory - 128GB is insufficient for the indexes without scalar quantization.

\subsection{Slot Filling Evaluation}

As an initial experiment we tried RAG with its default index of Wikipedia, distributed through Hugging Face. We refer to this as RAG-KKS, or RAG without the KILT Knowledge Source, as reported in Table \ref{tbl.dev}. Since the passages returned are not aligned to the KILT provenance ground truth, we do not report retrieval metrics for this experiment.  

Motivated by the low retrieval performance reported for the RAG baseline by \citet{kilt}, we experimented with replacing the DPR retrieval with simple BM25 (RAG+BM25) over the KILT knowledge source. We provide the raw BM25 scores for the passages to the RAG model, to weight their impact in generation. We also experimented with the Natural Questions trained DPR, with only RAG training on KILT (RAG+DPR$_{NQ}$). 

We use the approach explained in Section \ref{sec.approach} to train both the DPR and RAG models. \kgi{0} is a version of our system using DPR with hard negative samples from BM25.  The successor system, \kgi{1} incorporates DPR training using DNS.

The metrics we report include accuracy and F1 on the slot filler, where F1 is based on the recall and precision of the tokens in the answer, allowing for partial credit on slot fillers.  Our systems, except for RAG-KKS, also provide provenance information for the top answer.  R-Precision and Recall@5 measure the quality of this provenance against the KILT ground truth provenance.  Finally, KILT-Accuracy and KILT-F1 are combined metrics that measure the accuracy and F1 of the slot filler \textit{only when the correct provenance is provided}.

\begin{table}[t]
\begin{center}
\small
\begingroup
\renewcommand{\arraystretch}{1.2} 
\begin{tabular}{r|cccc}
& \textbf{R-Prec}  & \textbf{R@5} & \textbf{Acc.} & \textbf{F1} \\
\hline \multicolumn{5}{c}{\textbf{zsRE}} \\ \hline
RAG-KKS  &	 &  & 38.72 & 46.94\\
RAG+BM25 &	58.86 & 80.24 & 45.73 & 55.18  \\
RAG+DPR$_{NQ}$ & 68.13 & 79.19 & 46.03 & 55.75 \\
\kgi{0} & 96.24 & 97.53 & 69.58 & 77.24  \\
\kgi{1} & \best{98.60} & \best{99.70} & \best{71.32} & \best{78.85} \\	
\hline \multicolumn{5}{c}{\textbf{T-REx}} \\ \hline
RAG-KKS  &	 &  & 63.28 & 67.67   \\
RAG+BM25 &	46.40 & 67.31 & 69.10 & 73.11  \\
RAG+DPR$_{NQ}$ & 53.04 & 65.54 & 73.02 & 76.97 \\
\kgi{0} & 61.30 & 71.18 & 76.58 & 80.27 \\  
\kgi{1} & \best{74.34} & \best{82.89} & \best{84.04} & \best{86.89}
\end{tabular}
\endgroup
\end{center}
\caption{Dev sets performance for different retrieval methods}
\label{tbl.dev}
\end{table}

\begin{table*}[t]
\begin{center}
\small
\begingroup
\renewcommand{\arraystretch}{1.2} 
\begin{tabular}{r|ccccccc}
 & \textbf{R-Prec}  & \textbf{Recall@5} & \textbf{Accuracy} & \textbf{F1}  & \textbf{KILT-AC} & \textbf{KILT-F1}\\
\hline \multicolumn{7}{c}{\textbf{zsRE}} \\ \hline
\kgi{1} & \best{98.49} & \best{99.23} & \best{72.55} & \best{77.05} & \best{72.31} & \best{76.69} \\
\kgi{0}  & 94.18 & 95.19 & 68.97 & 74.47 & 68.32 & 73.45 \\
DensePhrases & 57.43 & 60.47 & 47.42 & 54.75 & 41.34 & 46.79 \\
GENRE & 95.81 & 97.83 & 0.02 & 2.10 & 0.00 & 1.85 \\
Multi-DPR & 80.91 & 93.05 & 57.95 & 63.75 & 50.64 & 55.44 \\
RAG {\small (KILT organizers)} & 53.73 & 59.52 & 44.74 & 49.95 & 36.83 & 39.91 \\
BART$_{LARGE}$ & N/A & N/A & 9.14 & 12.21 & N/A & N/A \\
\hline \multicolumn{7}{c}{\textbf{T-REx}} \\ \hline
\kgi{1}  & 74.36  & 83.14 & \best{84.36} & \best{87.24} & \best{69.14} & \best{70.58} \\
\kgi{0}  & 59.70 & 70.38 & 77.90 & 81.31 & 55.54 & 56.79 \\
DensePhrases & 37.62 & 40.07 & 53.90 & 61.74 & 27.84 & 32.34 \\
GENRE &	\best{79.42} & \best{85.33} & 0.10 & 7.67 & 0.04 & 6.66 \\
Multi-DPR & 69.46 & 83.88 & 0.00 & 0.00 & 0.00 & 0.00 \\
RAG {\small (KILT organizers)} &	28.68 & 33.04 & 59.20 & 62.96 & 23.12 & 23.94 \\
BART$_{LARGE}$ & N/A & N/A & 45.06 & 49.24 & N/A & N/A
\end{tabular}
\endgroup
\end{center}
\caption{KILT leaderboard top systems performance on slot filling tasks}
\label{tbl.test}
\end{table*}

Table \ref{tbl.dev} reports an evaluation on the development set, while Table \ref{tbl.test} reports the test set performance of the top systems on the KILT leaderboard.  \kgi{0} and \kgi{1} are our systems, while DensePhrases, GENRE, Multi-DPR, RAG for KILT and BART$_{LARGE}$ are explained briefly in Section \ref{sec.related}.  \kgi{1} gains dramatically in slot filling accuracy over the previous best systems, with gains of over 14 percentage points in zsRE and even more in T-REx.  The combined metrics of KILT-AC and KILT-F1 show even larger gains, suggesting that the \kgi{1} approach is effective at providing justifying evidence when generating the correct answer.  We achieve gains of 21 to 41 percentage points in KILT-AC.

Relative to Multi-DPR, we see the benefit of weighting passage importance by retrieval score and marginalizing over multiple generations, compared to the strategy of concatenating the top three passages and running a single sequence-to-sequence generation. 
GENRE is still best in retrieval for T-REx, suggesting that at least for a corpus such as Wikipedia, generating the title of the page can be very effective. 
A possible explanation for this behaviour is that most relations for a Wikipedia entity are mentioned in its corresponding page.

%% file: kilt_analysis2.tex
To explore the effect of retrieval on downstream performance we consider two variants of our systems: one using random passages from the index, forcing the system to depend on implicit knowledge, 
and the another using passages from the ground truth provenance, to measure the upper bound performance for the ideal retrieval system. 
Evaluation is reported in Table \ref{tbl.goldRandom} for 3 systems. 
By supplying these systems with the gold standard passages, we can see both the improvement possible through better retrieval, and the value of good retrieval during training. 
The best system, \kgi{1} is the most effective at generating slot fillers from relevant explicit knowledge because it was trained on more cases of justifying explicit knowledge. However, given random passages it is the worst. It has sacrificed some implicit knowledge for better capabilities in using explicit knowledge.

\begin{table}[t]
\small
\begin{center}
\begingroup
\renewcommand{\arraystretch}{1.2} 
\begin{tabular}{r|ccc}
Passages & RAG$_{\text{NQ}}$ & \kgi{0} & \kgi{1}\\
\hline
Retrieved & 70.58 & 76.58 & 84.04 \\
\hline
Gold &      88.66 & 89.46 & \best{90.20} \\
Random &    38.84 & \best{39.26} & \textit{36.64}
\end{tabular}
\endgroup
\end{center}
\caption{\label{tbl.goldRandom}T-REx Accuracy with Random and Gold Retrieval}
\end{table}




As shown in Table \ref{tbl.test}, BART$_{LARGE}$, which is the best implicit-knowledge baseline system for KILT slot filling, is approximately 40 points lower in in accuracy on T-REx if compared to \kgi{1}. To understand the impact of the explicit knowledge provided by DPR, we examine the improvement of \kgi{} over BART$_{LARGE}$. 
We consider two main hypotheses: 1) the value of explicit knowledge depends on the relation, and 2) the value of explicit knowledge depends on the corpus frequency of the entities related.

To evaluate hypothesis 1, we consider the most frequent 20 relations in the T-REx Dev set, each occurring at least 40 times. The relations with the lowest relative performance gain are taxonomy and partonomy relations: \relation{taxon-rank}, \relation{subclass-of}, \relation{instance-of}, \relation{part-of} and \relation{parent-taxon} as well as \relation{languages-spoken,-written-or-signed} and \relation{sport}. 
This suggest that essential properties of entities are well encoded in the language model itself. Inspecting the \relation{languages-spoken,-written-or-signed} we find that surface level information (i.e. French name vs. Russian name) is often sufficient for the correct prediction.  

In contrast, the relations that gain the most from explicit knowledge are: \relation{performer}, \relation{member-of-sports-team}, \relation{author}, \relation{place-of-birth}, \relation{country-of-origin}, \relation{cast-member}, \relation{director}. These relations are not central to the meaning of the head entity, like the taxonomy and partonomy relations, and are not typically predictable from surface-level features.

Regarding our second hypothesis, we might expect that more frequent entities have better representations in the parameters of a pre-trained language model, and that therefore the gain in performance due to use of explicit knowledge will show a strong dependence on the corpus frequency of the head or tail entity.

\begin{figure}[thb]
   \centering
   \includegraphics[width=0.95\linewidth]{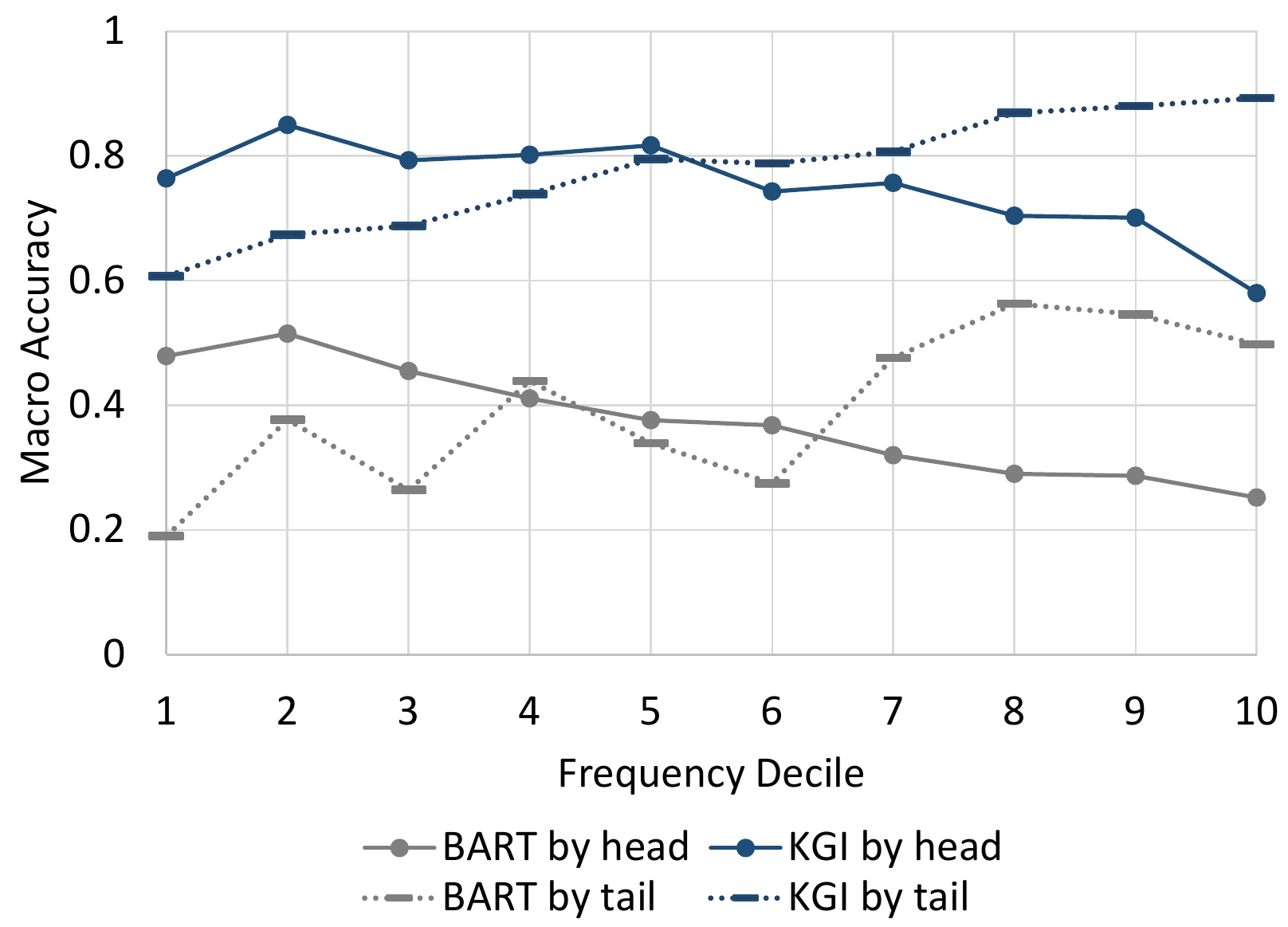}
   \caption{Performance as a function of entity frequency}
   \label{fig.entFreq}
\end{figure}

To test it, we group the Dev instances in T-Rex according to the decile of the head or tail entity frequency. 
We compute a macro-accuracy, weighting all relations equally. 
Figure \ref{fig.entFreq} shows the macro-accuracy of BART$_{LARGE}$ and \kgi{1} for each decile of head and tail entity frequency. Although there is a general trend of higher accuracy for more frequent tail entities and lower accuracy for more frequent head entities, \textit{there is no pattern to the gain of explicit knowledge over implicit knowledge from entity frequency}.  
There is a similar picture when considering the decile of the \emph{minimum} of the head or tail entity frequency. 
This falsifies our second hypothesis and suggests implicit knowledge is distinct in kind from explicit knowledge, rather than merely under-trained for low frequency entities.

